\documentclass[journal,twoside,web]{ieeecolor}
\usepackage{tmi}
\usepackage{cite}

\usepackage{amssymb, amsmath, amsfonts, mathrsfs, upgreek}
\usepackage{graphicx, epstopdf, tikz, color, pifont}
\usepackage{url, hyperref}
\usepackage{sidecap, rotating, multirow, subcaption}
\usetikzlibrary{matrix}
\usetikzlibrary{positioning}
\usepackage{rotating}

\usepackage[ruled, vlined, commentsnumbered, linesnumbered]{algorithm2e}
\usepackage{algpseudocode}

\newcommand{\ie}{{\em i.e.}}           

\newcommand{\bE}{\textbf{E}}
\newcommand{\bp}{\textbf{p}}
\newcommand{\bM}{\textbf{M}}
\newcommand{\bm}{\textbf{m}}
\newcommand{\bQ}{\textbf{Q}}
\newcommand{\bK}{\textbf{K}}
\newcommand{\bV}{\textbf{V}}
\newcommand{\bW}{\textbf{W}}
\newcommand{\bX}{\textbf{X}}
\newcommand{\bx}{\textbf{x}}

\newcommand{\bz}{\textbf{z}}
\newcommand{\bZ}{\textbf{Z}}
\newcommand{\ES}{\color{black}}

\def\BibTeX{{\rm B\kern-.05em{\sc i\kern-.025em b}\kern-.08em
    T\kern-.1667em\lower.7ex\hbox{E}\kern-.125emX}}
\begin{document}
\title{A Learnable Counter-condition Analysis Framework for Functional Connectivity-based Neurological Disorder Diagnosis}
\author{Eunsong Kang, Da-woon Heo, Jiwon Lee, and Heung-Il Suk \IEEEmembership{Senior Member, IEEE} 
\thanks{This work was supported by the Institute of Information \& Communications Technology Planning \& Evaluation (IITP) grant funded by the Korea government (MSIT) No. 2022-0-00959 ((Part 2) Few-Shot Learning of Causal Inference in Vision and Language for Decision Making) and No. 2019-0-00079, Artificial Intelligence Graduate School Program(Korea University). Also, this work was supported by the National Research Foundation of Korea (NRF) grant funded by the Korea government (No. 2022R1A4A1033856).}
\thanks{Eunsong Kang is with department of brain and cognitive engineering, Korea university, Seoul, 02841, Republic of Korea. (email: eunsong1210@korea.ac.kr)}
\thanks{Da-woon Heo and Jiwon Lee are with department of artificial intelligence, Korea university, Seoul, 02841, Republic of Korea.}
\thanks{Heung-Il Suk is with department of artificial intelligence and brain and cognitive engineering, Korea university, Seoul, 02841, Republic of Korea. (email: hisuk@korea.ac.kr)}}

\maketitle

\begin{abstract}
To understand the biological characteristics of neurological disorders with functional connectivity (FC), recent studies have widely utilized deep learning-based models to identify the disease and conducted post-hoc analyses via explainable models to discover disease-related biomarkers. Most existing frameworks consist of three stages, namely, feature selection, feature extraction for classification, and analysis, where each stage is implemented separately. However, if the results at each stage lack reliability, it can cause misdiagnosis and incorrect analysis in afterward stages. In this study, we propose a novel unified framework that systemically integrates diagnoses (\ie, feature selection and feature extraction) and explanations. Notably, we devised an adaptive attention network as a feature selection approach to identify individual-specific disease-related connections. We also propose a functional network relational encoder that summarizes the global topological properties of FC by learning the inter-network relations without pre-defined edges between functional networks. Last but not least, our framework provides a novel explanatory power for neuroscientific interpretation, also termed counter-condition analysis. We simulated the FC that reverses the diagnostic information (\ie, counter-condition FC): converting a normal brain to be abnormal and vice versa. We validated the effectiveness of our framework by using two large resting-state functional magnetic resonance imaging (fMRI) datasets, Autism Brain Imaging Data Exchange (ABIDE) and REST-meta-MDD, and demonstrated that our framework outperforms other competing methods for disease identification. Furthermore, we analyzed the disease-related neurological patterns based on counter-condition analysis. 
\end{abstract}


\section{Introduction}
\label{sec:introduction}
\IEEEPARstart{T}o comprehend the diverse pathological conditions of disorders such as autism spectrum disorder (ASD) and major depressive disorder (MDD), it is imperative to consider their neurological characteristics. In recent years, numerous functional MRI studies have identified neurological biomarkers associated with diseases. Functional connectivity (FC), temporal correlations of resting-state functional magnetic resonance imaging (rs-fMRI) signals among spatially distant brain regions, has been widely used in the literature \cite{kam2019deep, li2021virtual}.

Conventional brain disease studies have used voxel- and connection-wise statistical tests (\ie, independent $t$-test) to compare statistical differences between the patient and healthy control groups and determine the pathological regions or connections \cite{hong2019atypical, ciarrusta2020emerging, nair2020review}. Moreover, these statistical methods have been utilized as feature selection methods to improve the performance of the machine/deep learning classification models \cite{chen2019deriving, xu2020feature, nogay2020machine}. Concordantly, many neuroimaging studies also have used machine learning-based methods such as a least absolute shrinkage and selection operator (LASSO) and its variants for feature selection \cite{zille2017fused, li2019multimodal, li2020deep}. These methods have the advantage of eliminating features unrelated to the diagnosis. In general, a support vector machine (SVM) classifier has been implemented with the features chosen in the feature selection. 

However, independent implementation of feature selection and classifier may result in a suboptimal problem, which fails to provide the best predictive features for the classification model {\ES\cite{pudjihartono2022review}}. To address this concern in brain disease diagnosis, a method called recursive feature elimination (RFE) with SVM has been employed for disease diagnosis \cite{wang2019functional, dai2022alterations}. This technique iteratively eliminates a subset of features from the input data until the optimal set of features is obtained based on their importance evaluated by SVM weights. {\ES Furthermore, several deep models employ a feature selection using a learnable mask. For instance, a binary mask is thresholded by setting a quarter of its elements to one, and the remaining to zero \cite{wen2018neural}. Edge-mask learning for GCN models used the ReLU activation function to ensure the elements of the mask are non-negative and implemented $L_1$ regularization on the mask for its sparsity \cite{qu2021ensemble}. Despite the end-to-end process, these existing feature selection methods \cite{wang2019functional, dai2022alterations, wen2018neural, qu2021ensemble} primarily aim at identifying the connections that distinguish patients and normal controls. Accordingly, the features selected through these methods are applied across all patients, which may lead to the loss of individual-specific information.}


Although the most discriminative disease-related regions or connections are identified for a general understanding of the disease, elucidating the individual-specific difference in neurological patterns is also advantageous in clinical practice \cite{mueller2013individual}. {\ES However, it is challenging to consider the individual differences with group-based feature selection due to the substantial heterogeneity of the disease. This heterogeneity results from distinct neurological mechanisms among individuals and results in varied symptoms, even when diagnosed with the same disorder \cite{feczko2019heterogeneity}. To handle the heterogeneity of the disease in the feature selection stage,} a few recent studies have implemented a personalized ROI selection to capture individual variability in disease diagnosis. \cite{lee2021unified} devised a personalized ROI selection method by automatically selecting regions of interest (ROIs) via reinforcement learning (RL). Concretely, classification loss is used as the reward signal of RL for individual ROI selection. Furthermore, \cite{li2021braingnn} developed the ROI-based top-$k$ pooling to select salient ROIs {\ES for each individual during training.} Although personalized feature engineering methods can select features at the ROI level, they may not be able to account for the intricate topological properties of FC at the network level \cite{van2019cross}.


Along with considering personal traits, encoding FC relations for discriminative feature representations is also essential for diagnosis. Recently, many studies have proposed deep learning methods for discovering the topological features of FC \cite{kawahara2017brainnetcnn, li2021braingnn, li2021virtual}. FC is regarded as a graph structure where nodes and edges represent brain regions and their functional connections, respectively. These approaches demonstrated promising results in predicting behavior or cognitive states during the experiments. For example, BrainNetCNN \cite{kawahara2017brainnetcnn}, a convolutional neural network (CNN)-based model, encodes each row of FC with 1D convolution operations. Additionally, graph convolutional network (GCN)-based models have garnered attention for FC-based diagnoses. Most GCN-based models define ROIs as nodes and functional connectivities as edges. Node features usually consist of functional time series of ROI or ROI-based networks (each row of FC). One of the GCN-based models, BrainNetGNN \cite{li2021braingnn}, employs FC as a node feature and selects the top 10\% partial correlations as edges. By utilizing the ROI-aware graph convolutional layers and ROI pooling layer, BrainNetGNN demonstrates significant prediction performance for both disease diagnosis and task stimuli identification. Although these studies exhibited enhanced efficiency in prediction by considering the topological properties of FC, they are limited in their ability to cover the global context of FC owing to their high local dependencies. Specifically, the row-/column-wise 1D convolution of BrainNetCNN only works on connections within each seed-based network. Meanwhile, graph convolution in GCN-based models works on only pre-defined and strongly linked edges. Therefore, devising a model that explores the complex topological patterns of FC from a global perspective is desired.

Most conventional machine and deep learning classification models have adopted post-hoc explainable models such as a saliency map \cite{zhi2021bncpl}, which calculates the importance of input region/connection from the final output of the trained model. However, these methods have limited efficacy and reliability of the analysis \cite{kindermans2019reliability, arun2021assessing}. Specifically, additional modeling must be required after training the diagnostic model. Furthermore, these methods are vulnerable to noise{\ES\cite{adebayo2018sanity}}, and there is a wide diversity of opinions regarding malfunctioning regions{\ES\cite{van2012data, lau2019resting}}. Therefore, validating the involvement of these regions in disease identification and the extent to which they contribute to disease severity is challenging. To address these issues, an end-to-end framework for both classification and analysis that offers verifiable disease-related regions is desirable.

\begin{table}[t!]
	\centering
	\caption{Demographic characteristics of participants for ABIDE and REST-meta-MDD. The $p$-values from two-sample $t$-tests (Age, IQ, and Education) and Chi-squared test (Sex) between patients (\ie, ASD and MDD) and normal control (\ie, TD and HC) datasets are reported. STD: standard deviation}
	\renewcommand{\arraystretch}{1.2}
	\footnotesize{
	\begin{tabular}{cccc}\hline\hline
		& ASD            & TD             & $p$-value          \\ \hline
		Number of subjects     & 405            & 468            & -                \\
		Sex (Male/Female)    & 351/54         & 378/90         & 0.6980           \\
		Age (Mean/STD)       & 17.04$\pm$7.94   & 16.84$\pm$7.23   & 0.1325           \\
		IQ (Mean/STD)        & 105.81$\pm$17.06 & 111.09$\pm$12.06 & \textless 0.0001 \\ \hline\hline   & MDD            & HC             & $p$-value          \\\hline
		Number of subjects     & 830            & 771            & -                \\
		Sex (Male/Female)    & 306/524      & 316/455      & 0.1014           \\
		Age (Mean/STD)       & 34.39$\pm$11.57  & 34.55$\pm$13.14  & 0.8025           \\
		Education (Mean/STD) & 11.94$\pm$3.36   & 13.55$\pm$3.42   & \textless 0.0001 \\ \hline\hline
	\end{tabular}}
 \label{tab:demographic}
\end{table}

\begin{figure*}[!ht]
\centerline{\includegraphics[width=1.8\columnwidth]{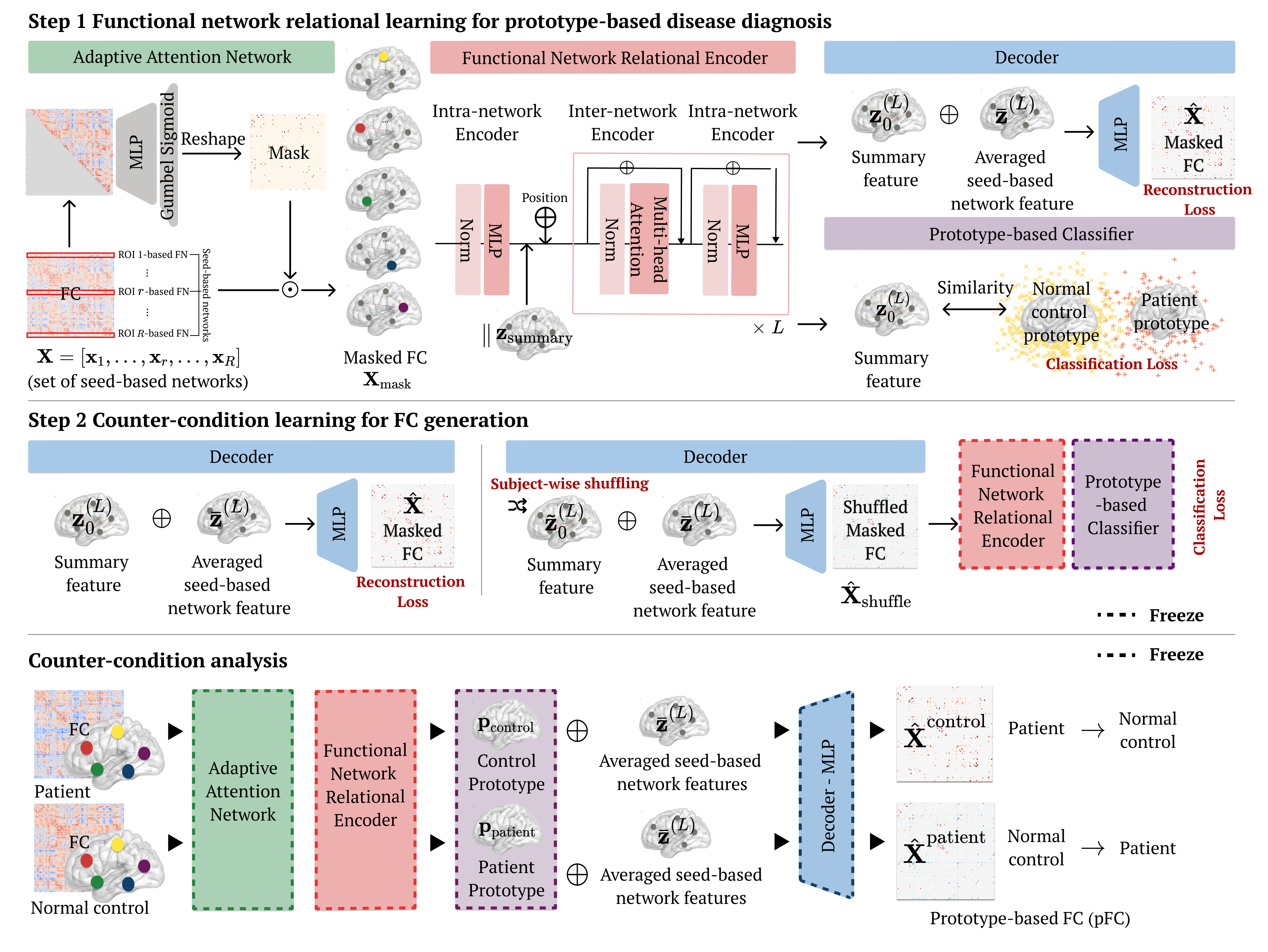}}
\caption{An overview of the proposed framework: two learning steps and counter-condition analysis. Our framework comprises four main components: an adaptive attention network, functional network relational encoder, decoder, and prototype-based classifier.}
\label{fig:overview}
\end{figure*}

\subsection{Contributions} 
In this study, we propose a unified framework that selects personal features at the connection level, summarizes the global context of FC for diagnosis, and provides reliable neuroscientific explanations. The four main contributions of this work can thus be summarized as follows: (1) We propose an adaptive attention network that can re-weigh the importance of connection features. (2) We devised a functional network relational learning method that summarizes inter-network functional connectivity as disease-related features. (3) Our model captures group-level disease-related characteristics, also termed prototypes (such as patient and normal control prototypes), and utilizes them for classification and elucidation. (4) Unlike the commonly adopted post-hoc analysis, our model is inherently interpretable. The explanation from our model starts with the question,``Where and how much FC should be changed?.” Notably, our model simulates a counter-condition of FC, \ie, alterations when a healthy individual is affected by the disease or the conditions under which patients with the disease may not be affected by the disease. We termed this approach \textit{counter-condition analysis}.

A preliminary version of this work \cite{kang2022prototype} was presented at the 2022 International Conference on Medical Image Computing and Computer-Assisted Interventions. This study extends the preliminary version with four main improvements: (1) designing novel adaptive feature selection networks, (2) devising an additional training step and a novel regularization term for a counter-condition explanation, (3) providing comprehensive qualitative and quantitative counter-condition analysis, and (4) demonstrating the applicability of the framework by validating it on an additional dataset, REST-meta-MDD.

\section{Dataset and preprocessing}
\label{sec:data}

{\ES We used rs-fMRI data from two publicly available datasets; Autism Imaging Data Exchange (ABIDE) I and the REST-meta-MDD\cite{chen2022direct}.} ABIDE I dataset provides image-based data of 1,112 individuals from 17 international organizations.  We excluded individuals with no files, those who had all zero values in specific ROIs, and those who failed the quality-control test conducted by three experts. Eventually, we used data from 873 individuals, which included 405 ASD patients and 468 individuals with typical development (TD). REST-meta-MDD dataset provides image-based data of 2,428 individuals from 17 hospitals in China. We removed the data by following the standard quality control procedures outlined in \cite{yan2019reduced}. We also excluded data from Site 4, as it was found to be a duplicate of Site 14. Thereby, we used 1,601 individuals, which included 830 MDD patients and 771 healthy control participants. {\ES The demographic details of participants from the ABIDE and REST-meta-MDD studies are provided in the Table \ref{tab:demographic}.}

{\ES In our study, both datasets were preprocessed with a Data Processing Assistant for Resting-State fMRI (DPARSF)\footnote{Scan procedure and protocols for ABIDE can be found at \url{http://fcon\textunderscore1000.projects.nitrc.org/indi/abide/}.} \footnote{Scan procedure and protocols for REST-meta-MDD can be found at \url{http://rfmri.org/REST-meta-MDD}} without global signal regression. We also applied ComBat harmonization \cite{fortin2017harmonization} to control the site-related effects. The brain was parcellated into 200 ROIs using Craddock (CC) atlas \cite{craddock2012whole}, a functionally-defined parcellation approach.} For both datasets, we calculated Pearson’s correlation coefficients to estimate FC using parcellated time-series fMRI data. Additionally, we initialized the diagonal elements in the FC matrix to zero.

\section{Proposed Method}
\label{sec:method}

\begin{table}[t]
\renewcommand{\arraystretch}{1.18} 
    \centering
        \caption{Summary of the notation}
\begin{tabular}{cl}\hline\hline
         Notation& \multicolumn{1}{c}{Description}\\\hline
         $\bX$& Input FC\\
        $\bM$& Mask\\
         $\bX_{\text{mask}}$& Masked FC\\
         $\hat{\bX}$& Reconstructed FC\\
         $\hat{\bX}_{\text{shuffle}}$& Simulated FC of counter-condition\\
 $\hat{\bX}^{c}$&Prototype-guided (simulated) FC\\
 $\bZ^{(l)}$&The output of $l$-th functional network relation encoder\\
$\bz_{\text{summary}}$
 & The summary feature\\
 $\bar{\mathbf{z}}$&Averaged seed-based network features\\
 $\bp^c$&Prototype features\\
 c & The class index (\ie, patient and control)\\\hline\hline
    \end{tabular}
    \label{tab:notation}
\end{table}

\subsection{Architecture overview}
Our framework comprises two learning steps and four main components, as illustrated in Fig. \ref{fig:overview}. The first step plays a major role in classification, whereas the second step aims to generate simulated FC. Our proposed model comprises an adaptive attention network, functional network relation encoder, prototype-based classifier, and decoder. The adaptive attention network emphasizes the significant connections of diagnosis that are fitted to each individual. Subsequently, the functional network relation encoder learns both local and global representations, that is, intra- and inter-network relations. Given the summary feature containing class-relevant information, the prototype-based classifier calculates the similarity between group-wise (\ie, ASD and TD, MDD and HC) prototypes for diagnosis. Meanwhile, the decoder encourages learning expressive features by reconstruction as well as generates FC in the opposite condition.

\subsection{Adaptive Attention Network}
Inspired by the scaling attention approach \cite{banville2022robust}, we propose an adaptive attention network to reduce the influence of redundant and insignificant connections on classification. Our adaptive attention network generates different attention masks conditioned on the functional connections of each individual. The attention mask reweights the connections according to the classification task on account of the end-to-end framework.

Given FC $\bX \in \mathbb{R}^{R \times R}$, where $R$ is the number of ROIs, we vectorize the upper triangular parts of the FC without diagonal elements, $\bx^{\text{upper}} \in \mathbb{R}^{\frac{R\times(R-1)}{2}}$. Subsequently, the adaptive attention network encodes functional connections into attention masks $\bm$ that are the same size as the input, as follows,
\begin{equation}    
	\begin{aligned}
		\bm^{\text{upper}} &= \text{Adaptive-Attention}({\bx^{\text{upper}}}) \\
		&= \text{Gumbel-sigmoid}(\bW_2(\text{ReLU}(\bW_1\bx^{\text{upper}}+\textbf{b}_1))+\textbf{b}_2)),
	\end{aligned}
\end{equation}
where the adaptive attention network has weight and bias parameters $\{\bW_1, \bW_2, \textbf{b}_1, \textbf{b}_2\}$, and two activation functions, ReLU and Gumbel-sigmoid. Notably, we used the Gumbel-sigmoid function \cite{geng2020selective} in the last layer to approximate the sampling process of discrete data. Thus, a subset of significant connections can be obtained. While the sigmoid function outputs values between zero and one, the Gumbel-sigmoid function maps inputs to either zero or one. Using the temperature $\tau$, we can relax this discrete property to be more deterministic. Unlike the feature selection methods conducted prior to training, our network can choose disease-related connections in a unified framework.

After reshaping  $\bm^{\text{upper}}$ to the upper and lower triangular parts of the FC matrix $\bM$, we masked the original FC with attention mask $\bM$ by element-wise multiplication, as follows, 
\begin{equation}
	\bX_{\text{mask}} = \bM \odot \bX.
\end{equation}
High masking scores maintained existing FC values, whereas low masking scores weakened the values to near zero. In other words, the mask emphasizes class-relevant connections by paying less attention to insignificant and redundant connections. Finally, the scaled FC with the attention mask $\bX_{\text{mask}}$ represents the connection that requires more focus for accurate diagnosis. Unlike the feature selection methods that apply to all individuals equally \cite{zhang2022detection, abrol2021deep, du2018classification}, $\bM$ varies with the functional properties of individuals, such that each individual has different subsets of attentive features.

\subsection{Functional Network Relation Encoder}
{\ES Our functional network relation encoder is designed to capture both local and global contexts of FC, motivated by \cite{dosovitskiy2021an}. Each row in the FC matrix is termed a seed-based (ROI-based) network, which captures the temporal correlations between a designated seed ROI and other ROIs. Consequently, FC can be considered a set of seed-based networks, consisting of $R$ number of seed-based networks. From the perspective of seed-based networks, intra-network relational learning focuses on understanding the relationships within each seed-based network. On the other hand, to learn interactions between seed-based networks is referred to as inter-network relational learning. While intra-network relational learning concentrates on the local aspects of FC, inter-network relational learning encompasses the broader and global aspects of FC.}

\subsubsection{Intra-network Relational Learning}
Given the masked FC as a sequence of seed-based networks, each row of ${\bX}_{\text{mask}}$, seed-based network, represents the temporal correlations of $r$-th ROI and the other ROIs (\ie, intra-network). First, we embed the intra-network relations of each masked seed-based network, as follows,
\begin{equation}
    \label{eq:intra}
    \begin{aligned}
    \bZ &= \text{Intra-network-Encoder}(\bX_{\text{mask}}) \\
    &= \text{MLP}(\text{LN}(\bX_{\text{mask}})),
    \end{aligned}
\end{equation}
{\ES where multi-layer perceptron (MLP) layers include two feed-forward layers and a GELU activation function. The MLP is shared across seed-based network features. Layer normalization (LN) \cite{ba2016layer} is applied to each seed-based network feature before the MLP layers. The Intra-network-encoder is designed such that its output preserves the dimensionality of the input, resulting in $\bZ \in \mathbb{R}^{R \times D}$.}

\subsubsection{Inter-network Relational Learning}
{\ES Before learning inter-network relations between seed-based networks}, we first included a learnable vector $\bz_{\text{summary}} \in \mathbb{R}^{1 \times R}$ to the set of seed-based network features, $\bZ$. Furthermore, a sinusoidal positional encoding matrix, $\bE \in \mathbb{R}^{(R+1) \times D}$, was added for relative regional information. Therefore, the inter-network encoder accepts $\bZ^{0}$ as the input,
\begin{equation}
	\bZ^0 = [\bz_{\text{summary}}, \bz_1, ..., \bz_R] + \bE.
\end{equation}
The inter-network encoder learns interactions between seed-based networks via multi-head self-attention (MHSA) \cite{vaswani2017attention}. Particularly, the self-attention (SA) mechanism is essential for learning inter-network relations:
\begin{equation}
	\label{eq:sa}
	\text{SA}(\cdot) = \text{Self-Attention}(\bQ, \bK, \bV) = \text{Softmax}(\frac{\bQ\bK^{\top}}{\sqrt{D}})\bV,
\end{equation}
 where $\bQ, \bK$, and $\bV\in\mathbb{R}^{(R+1)\times D}$ are referred to as the query, key, and value, respectively. The query, key, and value represent the linear transformation of input $\bZ^{0}$ with learnable weight parameters $\bW_Q, \bW_K$, and $\bW_V$, thus, $\bQ=\bZ \bW_Q, \bK=\bZ \bW_K$, and $\bV=\bZ \bW_V$ with bias, respectively. Specifically, the dot product of $\bQ$ and $\bK$ measures the similarity between seed-based networks, and the softmax function normalizes these similarities. By multiplying $\bV$, the output of self-attention is proportional to the inter-network similarity score. With this module, our proposed network can encode the relations of network pairs to represent an inter-network graph.

Rather than a single SA for single inter-network relations, we jointly encode multiple representations with a multi-head SA to learn various relational patterns, 
\begin{equation}
\begin{aligned}    
	\text{MHSA}(\cdot) &= \text{Multi-head-SA}(\bQ, \bK, \bV)\\ &= \text{Concat}(\text{SA}_1, \text{SA}_2, ..., \text{SA}_H)\bW_{\text{MHSA}},
 \end{aligned}
\end{equation}
where $H$ is the number of heads. Each single-head SA is calculated by its respective $\bQ_h, \bK_h$, and $\bV_h \in \mathbb{R}^{(R+1)\times(D/H)}$. Subsequently, the outputs of the single-head SA are aggregated and linearly transformed by $\bW_{\text{MHSA}} \in \mathbb{R}^{D \times D}$. Therefore, the inter-network encoder is defined as follows: 
\begin{equation}
    \text{Inter-network-Encoder}(\bZ^0) = \text{MHSA}(\text{LN}(\bZ^0)).
\end{equation}

\subsubsection{Functional Network Relation Encoder}
{\ES Taken together, each encoder block consists of two types of encoders, intra-network encoder and inter-network encoder, as follows, 
\begin{equation}
    \bZ' = \text{Inter-Network-Encoder}(\bZ^{(l-1)}) + \bZ^{(l-1)}
\end{equation}
\begin{equation}
\label{eq:tfmlp}
    \bZ^{(l)} =\text{Intra-Network-Encoder}(\bZ') + \bZ',
\end{equation}
where $l = \{1,  ..., L\}$ represents the number of blocks. The residual connections are implemented after the Inter-Network-Encoder and Intra-Network-Encoder. As the summary vector is employed for classification, it captures diagnostic information by iteratively learning the local and global patterns of FC.}

\subsection{Prototype-based Classifier}
For classification, the similarity is calculated between the summary feature vector in the last layer and learnable class prototypes $\bp_c \in \mathbb{R}^{1 \times D}$, where $c=\{\text{patient}, \text{control}\}$. The prototypes are representative vectors of each class, which are the centroids of the summary features of each class. These were randomly initialized and trained using prototype learning \cite{yang2018robust}. The probability $p(c|$\bX$)$ is derived as follows:
\begin{equation}
\label{eq:probforce}
      p(c|\bX) = \frac{\exp^{s(\bz_{\text{summary}}, \bp_{c})}}{\sum_{i \in C}\exp^{s(\bz_{\text{summary}}, \bp_{i})}},
\end{equation}
where $c$ represents a class index. We used cosine similarity for the similarity measure $s$ and adopted cross-entropy for the classification loss $\mathcal{L}_{\text{class}}$. 

The label of the individual summary feature vector is predicted to be the prototype(s) index with the highest similarity. {\ES More explicitly, an input instance showing high similarity to the patient prototype is identified as a patient.} Simultaneously, the prototypes are trained to be class-representative by increasing their similarity with the assigned summary feature vector. {\ES In scenarios with more than two classes, the number of prototypes augments in proportion to the number of classes. Consequently, this prototype-based classifier enables the extension of binary classification to multi-class classification.}

\begin{table*}[h!]
\centering
\caption{Average classification results (mean$\pm$standard deviation) for ASD and MDD identification. The highest performance was highlighted for each evaluation metric. * denotes statistical significance ($p$$<$0.05).}
\renewcommand{\arraystretch}{1.18} 
\begin{tabular}{lllllllll}\hline\hline
                                     & \multicolumn{4}{c}{\textbf{ABIDE}}                                                                                                                            & \multicolumn{4}{c}{\textbf{REST-meta-MDD}} \\\hline
\multicolumn{1}{c}{\textbf{Methods}} &\multicolumn{1}{c}{\textbf{AUC}} & \multicolumn{1}{c}{\textbf{ACC (\%)}} & \multicolumn{1}{c}{\textbf{SEN (\%)}} & \multicolumn{1}{c}{\textbf{SPC (\%)}} & \multicolumn{1}{c}{\textbf{AUC}} & \multicolumn{1}{c}{\textbf{ACC (\%)}}             & \multicolumn{1}{c}{\textbf{SEN (\%)}}             & \multicolumn{1}{c}{\textbf{SPC (\%)}} \\\hline
BrainCNN \cite{kawahara2017brainnetcnn}                      & 0.6536$\pm$0.02* & 61.02$\pm$0.01*                        & 60.37$\pm$0.03*                        & 61.59$\pm$0.03*                        &0.7129$\pm$0.01*   &65.71$\pm$0.50*  & 65.01$\pm$1.63*                & 66.46$\pm$1.45*    \\
BrainCNN-P \cite{zhi2021bncpl}                          & 0.6668$\pm$0.06* & 62.21$\pm$0.08*         & 62.45$\pm$1.76*         & 62.00$\pm$2.39*         & 0.7011$\pm$0.02*   & 64.39$\pm$2.38* & 65.04$\pm$4.35* & 63.70$\pm$6.23*    \\
GCN \cite{kipf2016semi}                                 & 0.6425$\pm$0.01* &	64.32$\pm$1.12* & 63.22$\pm$1.65*	& 65.28$\pm$1.99*& 0.6372$\pm$0.01* &	63.85$\pm$0.86*	& 65.38$\pm$1.17*&	62.05$\pm$2.15*\\
GAT \cite{velivckovic2017graph}                                 &  0.6568$\pm$0.03* & 65.63$\pm$2.56*	& \textbf{66.45}$\pm$3.40 &	64.92$\pm$2.93* &	0.6808$\pm$0.02*	& 68.10$\pm$1.40* & 68.21$\pm$1.94	& 67.94$\pm$4.17*          \\
GraphSage \cite{hamilton2017inductive}                           &    0.6406$\pm$0.01* &	64.11$\pm$1.12*	& 63.51$\pm$1.79*	& 64.61$\pm$3.18*&	0.6557$\pm$0.02*	& 65.57$\pm$1.74*	& 65.65$\pm$3.62*	& 65.48$\pm$3.12*\\
STCAL \cite{liu2023spatial}                      & 0.6918$\pm$0.01      & 63.97$\pm$0.80*   & 61.44$\pm$2.47*                 & 65.01$\pm$1.90*
              & 0.7331$\pm$0.01         & 66.98$\pm$0.70*            & \textbf{70.24}$\pm$1.22          & 63.32$\pm$1.53*
                                      \\
STAGIN (GARO) \cite{kim2021learning}                     & 0.5851$\pm$0.01*            & 55.66$\pm$0.12*                 & 58.20$\pm$0.38*                                      & 55.93$\pm$0.11*                                       &0.7096$\pm$0.01*                                       &66.56$\pm$0.34*                                                   &68.03$\pm$2.21*                                                   &64.57$\pm$3.05*                                       \\
STAGIN (SERO) \cite{kim2021learning}                     & 0.6046$\pm$0.01*           & 58.81$\pm$0.18*                & 57.75$\pm$4.21*                                       & 60.00$\pm$3.96*                                       &0.6904$\pm$0.01*                                       &65.29$\pm$0.90*                                                   &64.21$\pm$0.69*                                                   &65.50$\pm$2.00*                                       \\\hline
$t$-test + SVM                       & 0.6875$\pm$0.01*   & 63.93$\pm$1.13*   & 60.19$\pm$2.01*    & 67.16$\pm$2.05*    & 0.6789$\pm$0.01*                       & 63.15$\pm$0.87*                                    & 65.01$\pm$1.55*                                    & 61.15$\pm$1.95*                        \\
LASSO + SVM \cite{tibshirani1996regression}                         & 0.6816$\pm$0.01*   & 64.07$\pm$2.09*    & 60.53$\pm$2.91*  & 67.05$\pm$1.69*    & 0.7354$\pm$0.01                       & 67.72$\pm$0.95*                                    & 69.35$\pm$0.67                                    & 66.02$\pm$1.77*                        \\
RFE-SVM \cite{guyon2002gene}                             & 0.6812$\pm$0.10*   & 63.95$\pm$1.58*    & 59.66$\pm$1.60*    & 66.11$\pm$2.65*    & 0.7296$\pm$0.01*   & 67.81$\pm$0.80*          & 70.01$\pm$1.31                & 65.50$\pm$1.57*    \\
BrainGNN \cite{li2021braingnn}                            & 0.6113$\pm$0.13*   & 59.97$\pm$1.80*    & 54.65$\pm$4.03*    & 64.51$\pm$5.71*    & 0.6151$\pm$0.01*&	60.36$\pm$0.65*&	62.24$\pm$4.99* & 56.17$\pm$6.32*          \\
t-test + P                    & 0.6579$\pm$0.01*   & 63.67$\pm$0.09*    & 61.91$\pm$3.29*    & 65.20$\pm$2.77*    & 0.6736$\pm$0.01*   & 65.11$\pm$0.40*                & 63.01$\pm$1.89*                & 67.19$\pm$2.58*    \\
LASSO + P \cite{tibshirani1996regression}                    & 0.6539$\pm$0.01*  & 63.44$\pm$0.08*    & 63.25$\pm$3.80*    & 63.55$\pm$3.99*    & 0.6729$\pm$0.01*   & 65.15$\pm$0.38*               & 65.24$\pm$1.21*               & 64.77$\pm$1.01*    \\
Binary Mask + P \cite{wen2018neural}                          & 0.6429$\pm$0.01* &	61.88$\pm$0.58*	& 61.94$\pm$2.74*&	61.82$\pm$1.58*	& 0.6684$\pm$0.01*	& 64.00$\pm$0.44*	& 65.89$\pm$4.16*	& 61.93$\pm$3.42* \\
Edge Mask + P \cite{qu2021ensemble}                           & 0.6430$\pm$0.01* &	62.57$\pm$0.52*	& 60.19$\pm$2.82* &	64.60$\pm$2.54* &	0.6706$\pm$0.01*	& 63.36$\pm$0.28*	& 63.16$\pm$1.04*	& 63.42$\pm$1.17*                            \\
\hline
Preliminary \cite{kang2022prototype} & 0.6869$\pm$0.01*& 65.00$\pm$1.34* &	63.05$\pm$3.61* & 66.61$\pm$3.39* & 0.7053$\pm$0.01* &	68.21$\pm$0.72* & 68.95$\pm$1.47	& 67.08$\pm$2.72* \\
P w/o $\bM$ & 0.6565$\pm$0.01* &	63.46$\pm$1.98* & 62.82$\pm$4.16* & 65.42$\pm$1.15* & 0.6874$\pm$0.01* & 66.42$\pm$1.23* & 68.00$\pm$1.56* & 65.58$\pm$0.94*\\
P w/o intra-network        & 0.6962$\pm$0.02 & 64.12$\pm$1.97*& 62.29$\pm$5.58*	& 69.50$\pm$2.55 & 0.7145$\pm$0.01* & 67.80$\pm$0.16* & 69.03$\pm$2.86 & 65.24$\pm$4.11* \\
P w/o prototype & 0.6682$\pm$0.01* & 62.91$\pm$0.94* & 64.02$\pm$2.35 & 61.83$\pm$2.52* & 0.7300$\pm$0.01*& 67.72$\pm$0.98* & 69.34$\pm$0.71 & 62.64$\pm$1.86* \\\hline
Proposed                             & \textbf{0.7150}$\pm$0.01                       & \textbf{67.91}$\pm$0.91                        & 65.45$\pm$3.93                        & \textbf{70.41}$\pm$3.32                        & \textbf{0.7562}$\pm$0.01                       & \textbf{70.19}$\pm$0.83                                    & 69.70$\pm$2.69                                    & \textbf{70.68}$\pm$2.11         \\\hline\hline              
\end{tabular}
\label{tab:result1}
\end{table*}

\subsection{Decoder}
We introduced three decoding approaches using the same decoder by varying the input combinations. The decoder was trained by reconstruction  (Eq. \ref{eq:decrec} and Eq. \ref{eq:decrec2}) in Step 1 and counter-condition learning (Eq. \ref{eq:probforce} and Eq. \ref{eq:deccl}) in Step 2. The decoder in Eq. \ref{eq:11} is not involved in training but is used for counter-condition analysis.

\subsubsection{Decoder for Reconstruction} 
We reconstruct the FC in Step 1 as follows,
\begin{equation}
	\label{eq:decrec}
	\hat{\bX} = \text{Decoder}(\bar{\bz} + \bz_{\text{summary}}).
\end{equation}
where $\hat{\bX}$ is the reconstructed FC and the feature $\bar{\bz}$ is the average of $R$ number of seed-based networks embeddings, $\bar{\bz} = \frac{\sum^{R}_{i=1}\bZ^{(L)}_i}{R}$, {\ES for each instance}. The decoder is defined by two fully connected layers with a nonlinear activation function. Considering that the summary feature $\bz_{\text{summary}}$ is utilized for classification, it involves information that distinguishes between patients and normal controls, thereby regarded as a group-level representation. On the other hand, the averaged feature $\bar{\bz}$ is an embedding of FC, which reflects individual-level characteristics. We utilized the $L_1$ loss for reconstruction: 
\begin{equation}
		\label{eq:decrec2}
	\mathcal{L}_{\text{recon}} = \|\hat{\bX} - \bX_{\text{mask}}\|.
\end{equation}
We reconstructed $\bX_{\text{mask}}$ instead of $\bX$ only to focus on the connections selected in the adaptive attention network to better fit the prototype-based classifier.

\subsubsection{Decoder for Counter-condition Learning}
\label{sec:counterlearning}
To allow the decoder to generate a counter-condition FC in Step 2, we first replaced the summary feature vector with that of a sample in the other class. Specifically, we shuffled the summary features in each mini-batch to replace a subject's class information with the opposite class information of others. Consequently, the decoder can generate a masked FC of the opposite class,
\begin{equation}
	\label{eq:deccl}
	\hat{\bX}_{\text{shuffle}} = \text{Decoder}(\bar{\bz} + \text{Shuffle}(\bz_{\text{summary}})),
\end{equation}
{\ES where $\text{Shuffle}$ denotes a permutation function.} Subsequently, the simulated FC $\hat{\bX}_{\text{shuffle}}$ was passed through the functional network relational encoder and the prototype-based classifier. {\ES Due to the changes in the class attributes, referred to as the counter-condition, it should be classified as belonging to the opposite class in the classification task.} To differentiate the classification that utilizes the original FC, the classification with simulated counter-condition FC is termed counter-condition classification.

\subsubsection{Decoder for Counter-condition Analysis}
\label{sec:decforcountercondition}
Similar to the shuffling of the summary feature vector, we can also substitute the summary feature vector of an individual into a representative prototype vector. Subsequently, we decoded seed-based network features of an individual with the group prototype to produce prototype-guided FC (pFC),  $\hat{\bX}^{c}$, as follows, 
\begin{equation}
	\label{eq:11}
	\hat{\bX}^{c}= \text{Decoder}(\bar{\bz} + \bp_c),
\end{equation}
where $c$ represents a class index, patient or normal control. To maintain the same influence of the summary feature vector, we normalized $\bp_c$ to the same magnitude as the original summary features. 

Consequently, the pFC maintains individual characteristics {\ES inherent to the averaged seed-based network  features}, whereas individual class-related characteristics are replaced with group-representative characteristics. Notably, replacing the summary feature vector of a healthy individual with the patient prototype $\bp_{\text{patient}}$ can help predict the functional degradation of the FC, assuming that the individual is affected by the disease. Meanwhile, for patients with the healthy prototype $ \bp_{\text{control}}$, we can identify the distinguishing variations of FC as if the individuals follow a normal development or cognitive process. {\ES Similarly to $\bar{\bX}_{\text{shuffle}}$, pFC that changes in the class attributes is categorized under counter-condition FC.} Our proposed method can simulate counter-condition FC for each individual, which is beneficial for obtaining deeper insights into the functional characteristics of the brain with the disease.

\subsubsection{Regularization}
Simultaneously, we specifically reduce the influence of seed-based network features as follows,
\begin{equation}
       \mathcal{L}_{\text{reg}} = \sum_{i \in C}|s(\bar{\bz}, \bp_{i})|.
\end{equation}
By minimizing the absolute value of the similarity between $\bar{\bz}$ and prototypes $\bp_c$, seed-based network features become less informed of class characteristics. Without regularization, the residual class information in seed-based network features negatively affects the generation of the FC of the opposite class.

\subsection{Overall Objective Function and Training Strategy}
Our framework has two steps: Step 1 for classification and reconstruction and Step 2 for counter-condition FC generation. The objective function of each step comprises different losses,
\begin{equation}
	\mathcal{L}_{\text{step}_1} = \mathcal{L}_{\text{recon}} + \mathcal{L}_{\text{class}} +\mathcal{L}_{\text{reg}},
\end{equation}
\begin{equation}
	\mathcal{L}_{\text{step}_2} = \lambda_{\text{recon}}\mathcal{L}_{\text{recon}} + \lambda_{\text{class}}\mathcal{L}_{\text{class}}, 
\end{equation}
where $\lambda_{\text{recon/class}}$ represents the hyperparameter that controls the balance of each loss. We control the balance between the reconstruction loss and the classification loss in  $\mathcal{L}_{\text{step}_2}$ to generate FC that contains information on the counter-condition with $\lambda_{\text{class}}$ while maintaining a subject's own functional configuration with $\lambda_{\text{recon}}$ to avoid losing reconstruction capacity.

Step 1 and Step 2 are trained by the turns of the epoch. The alternation of these two steps in the training process enables disease identification and counter-condition FC generation for analysis in a unified framework. In Step 1, all modules, \ie, adaptive attention network, functional network relational encoder, decoder, and prototype classifier, are trained. However, in Step 2, only the decoder is trained while the other networks are fixed. In this step, the classification loss of $\hat{\bX}_{\text{shuffle}}$ is minimized to guide the decoder to generate a counter-condition FC.

\section{Results}
\subsection{Comparative Methods}
\label{sec:comparative}
We demonstrate the efficacy and validity of our proposed method in two aspects: functional network relational learning and feature selection. From the perspective of functional network relational learning, we chose BrainNetCNN (BrainCNN) \cite{kawahara2017brainnetcnn}, designed to learn network features with a 1D convolution operation, for comparison. As our model uses a prototype-based classifier, we added a prototype-based BrainNetCNN that predicts the label with prototypes (BrainCNN-P) \cite{zhi2021bncpl}. We also compared our method with the representative GNN models: GCN \cite{kipf2016semi}, GAT \cite{velivckovic2017graph}, GraphSage \cite{hamilton2017inductive}, and BrainGNN \cite{li2021braingnn}. {\ES As our functional network relational learning is grounded in Transformer architecture, we compared our model with other Transformer-based models: spatio-temporal attention graph isomorphism network (STAGIN) \cite{kim2021learning} and spatial-temporal co-attention learning (STCAL) \cite{liu2023spatial}.}

We additionally adopted three {\ES traditional} feature selection methods: an independent $t$-test for each connection as a univariate feature selection, a LASSO \cite{tibshirani1996regression} and a RFE \cite{guyon2002gene} as a multivariate feature selection. These are group-level feature selections in which all samples have the same selected features. We selected significant functional connections via a $t$-test or LASSO and conducted a SVM classifier, a conventional diagnostic framework in neuroimaging. Meanwhile, RFE-SVM recursively ranked the features based on the SVM weights and removed the least informative and redundant features until only the desired number of features remains. Furthermore, we replaced an adaptive attention network with these traditional feature selection methods and trained the functional network relational encoder, decoder, and prototype-based classifier ($t$-test+P and LASSO+P in Table \ref{tab:result1}). {\ES The connections selected via the $t$-test and LASSO are consistent across all individuals, in contrast to our model which selects connections on an individual basis. Consequently, we engaged in a comparison between group-level and individual-level feature selection. 

Furthermore, we conducted a comparison between two deep learning-based feature selection methods and our adaptive attention mask: the binary mask method \cite{wen2018neural} and the edge mask \cite{qu2021ensemble}. It is worth noting that these masks are trainable within an end-to-end learning framework. For contrasting ROI-level and network-level feature selection, we selected BrainGNN \cite{li2021braingnn}, a notable GNN model, as a comparative model. While our model operates by selecting connections at the individual level, BrainGNN is especially adept at ROI selection at this level, facilitated by its pooling layer. We also included our prior model \cite{kang2022prototype} published in MICCAI (Preliminary), for comparison.} 



\subsection{Experimental Settings}
\label{sec:implementation}
We used the same hyperparameters for both datasets. Our adaptive attention network used two fully connected layers with hidden units $\{\lfloor\frac{R\times(R-1)}{4}\rceil, \allowbreak \lfloor\frac{R\times(R-1)}{2}\rceil\}$. {\ES The $\tau$ value of the Gumbel-sigmoid was set to 5.0 for both datasets. Functional network relational learning comprises a two-block ($L$=2) Transformer encoder, where the number of heads was 10 ($H$=10) and the hidden feature dimension of the encoder was 128.} The output feature dimension of the encoder was as same as the number of ROIs ($D$=$R$). The decoder was composed of two fully connected layers with hidden sizes of $\{\lfloor\frac{R\times(R-1)}{4}\rceil, \allowbreak \lfloor\frac{R\times(R-1)}{2}\rceil\}$. The cross-entropy loss with a softmax temperature of 0.5 is used for disease identification. All modules were optimized using the AdamW optimizer \cite{loshchilov2017decoupled}. The learning rate of Step 1 was 0.0005. In Step 2, only the decoder was trained with a learning rate of 0.0001. The batch size was 32 in all steps. To avoid losing the reconstruction ability of the decoder, the $\lambda_{\text{rec}}$ and $\lambda_{\text{cls}}$ in Step 2 were set to 1.0 and 0.1, respectively. To avoid overfitting, we applied $L_2$ regularization of 0.0001 and dropout with a drop rate of 0.5.

For competing methods, we followed the same hyperparameter setting as reported in the original manuscript for BrainCNN, BrainCNN-P, and BrainGNN. For GAT and Graphsage, where no fMRI data were used in the original manuscript but used as a comparative method in BrainGNN, the same hyperparameter settings in BrainGNN, except for the learning rate, were used. In addition to the original learning rate setting, we also identified the optimal learning rate among  $\{1e^{-2}, 1e^{-3}, 1e^{-4},  5\times 1e^{-2}, 5\times 1e^{-3}, 5\times 1e^{-4}\}$ by tuning the validation set. We selected connections with the $p$-value of less than 0.05 as a feature selection method, whereas the  $\lambda$ for LASSO was selected from $\{0.001, 0.002, ... ,0.01\}$. The $C$ parameter for SVM and {\ES $\beta$ of $L_1$ sparsity term in Edge Mask} were chosen from $\{1e^{-1}, 1e^{-2}, 1e^{-3}\}$. For RFE-SVM, we predefined the number of features following the previous research \cite{wang2019functional} for ABIDE and \cite{dai2022alterations} for REST-meta-MDD. We conducted five-fold cross-validation and repeated it five times for the proposed and all competing methods.

\subsection{Performance Comparison}
\label{sec:results}

\begin{table}[t!]
\caption{Average counter-condition classification results (mean$\pm$standard deviation). Ablation studies on Step 2 and regularization loss $\mathcal{L}_{\text{reg}}$ in the proposed model for counter-condition classification.}
\label{tab:result2}
\renewcommand{\arraystretch}{1.2}
\setlength{\tabcolsep}{3.7pt}	
\begin{tabular}{lcccc}\hline\hline
\multicolumn{5}{c}{ABIDE}                                 \\\hline
                         & AUC    & ACC (\%)   & SEN (\%)   & SPC (\%)   \\\hline
w/o Step 2               & 0.6876$\pm$0.06 & 59.69$\pm$10.20 & 74.49$\pm$18.80 & 44.87$\pm$6.81 \\
w/o $\mathcal{L}_{\text{reg}}$ & 0.9009$\pm$0.06 & 85.04$\pm$11.73 & 90.16$\pm$6.55 & 82.99$\pm$16.89  \\
Proposed                 & 0.9454$\pm$0.06 & 92.84$\pm$7.27 & 88.19$\pm$11.00 & 96.14$\pm$4.33\\\hline\hline

\multicolumn{5}{c}{REST-meta-MDD}  \\\hline
         & AUC    & ACC (\%)   & SEN (\%)   & SPC (\%)   \\\hline
w/o Step 2               & 0.6869$\pm$0.05 & 61.25$\pm$4.56 & 72.72$\pm$9.22 & 49.78$\pm$12.78 \\
w/o $\mathcal{L}_{\text{reg}}$ & 0.9313$\pm$0.03 & 91.31$\pm$6.15 & 92.16$\pm$6.71 & 93.13$\pm$3.85 \\
Proposed                 & 0.9707$\pm$0.02 & 94.61$\pm$1.77 & 98.51$\pm$1.90 & 90.66$\pm$2.67\\\hline\hline
\end{tabular}
\end{table}

The results are summarized in Table \ref{tab:result1}. We reported the mean and standard deviation of the classification performance using four metrics: area under the receiver operating characteristic curve (AUC), accuracy (ACC), sensitivity (SEN), and specificity (SPC). Our method achieved the best ACC of 67.91\% and a high AUC of 0.7150, SEN of 65.45\%, and SPC of 70.41\% on the ABIDE dataset. Additionally, our method obtained the best ACC of 70.19\%, AUC of 0.7562, and SPC of 70.68\% on the REST-meta-MDD dataset. In addition, we performed the Wilcoxon signed-rank test ($p<$0.05) to validate the superiority of our method.

Our model achieved significant improvements on two separate datasets in AUC compared with that using the graph-based models. This can be attributed to two factors. First, our model covers the entire brain relations ($R$$\times$$R$) owing to the self-attention mechanism, compared with that in BrainCNN and BrainCNN-P, which share the weights of 1$\times$$R$ convolution for all seed-based networks. Our model can capture subtle and globally distributed information in the FC. Second, compared with GNN-based models that predefine edges by thresholding (such as the top 10\% positive or absolute values of FC) prior to feature extraction, our model selects connections using adaptive attention networks within a unified framework. In addition, our functional network relational encoder learns various network relations via a multi-head self-attention operation.

{\ES We also performed evaluations in comparison to Transformer-based models, such as STAGIN and STCAL. These methods showed degraded classification performance in ACC and SPC, especially when applied to the ABIDE dataset. Despite their ability to capture both intra-network and inter-network relations of FC, their use of all connections without feature selection could result in a decline in model performance, as it could learn irrelevant or noisy patterns.}

The results of our study demonstrate that our proposed method outperforms competing methods that use conventional feature selection (such as $t$-test, LASSO, and RFE-SVM) and optimizable masks (such as Binary Mask and Edge Mask) on both datasets. These findings suggest that our end-to-end approach with personalized feature selection is more effective than group-wise feature selection conducted independently of the classifier. Despite a unified framework for selecting task-relevant ROIs, BrainGNN performed poorly when using rs-fMRI. This finding indicates that the connection-level feature selection approach outperforms the ROI-level feature selection approach. 


\subsection{Counter-condition Classification}

To evaluate the simulated counter-condition FC, we conducted counter-condition classification. The performance of the counter-condition classification is estimated in Table \ref{tab:result2}. As mentioned in Sec. \ref{sec:decforcountercondition}, we replaced the summary feature with the prototype of the counter-condition and reconstructed the {\ES pFC, FC with class information of the opposite condition, denoted as $\hat{\mathbf{X}}^{c}$, where $c$ represents a class index.} Using the pFC as an input, we identified the disease using a trained prototype-based classifier. The label is assigned the same index as that of the prototypes. Only correct test samples in classification were considered in the counter-condition classification.

Counterintuitively, in counter-condition classification, our proposed method exhibited high performance in all metrics. Notably, the counter-condition classification with pFC is not implemented in training phase. {\ES As our summary feature is trained by maximizing the similarity between the class prototypes and Step 2 where the decoder is conditioned to produce a counter-condition FC using others' summary features from the opposite class, these underlying mechanisms contribute to the performance of our method in counter-condition classification.}
\begin{figure}[t!]
\includegraphics[width=0.48\textwidth]{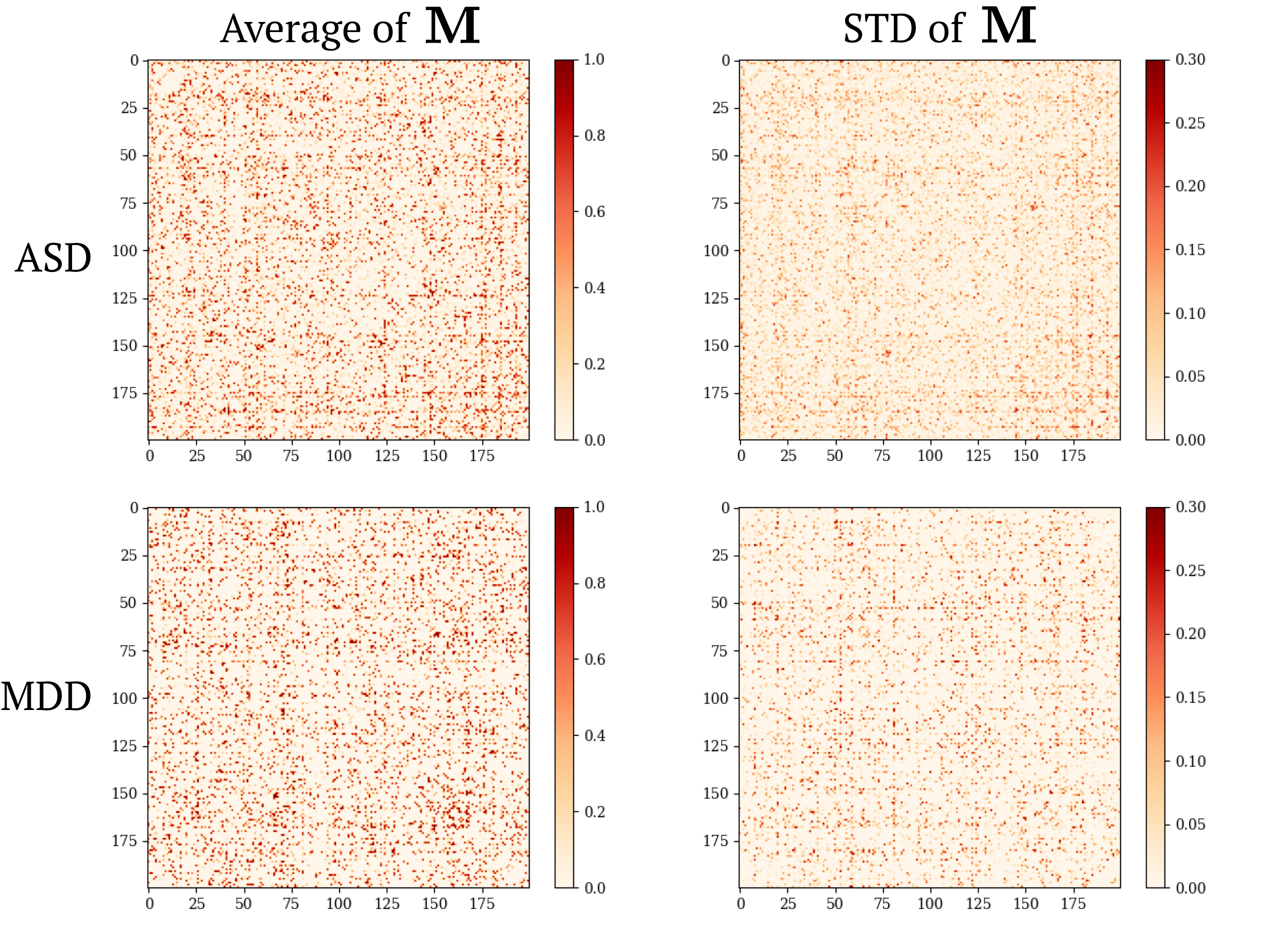}
\caption{Average and standard deviation (STD) of masks $\bM$ for ASD and MDD patients.}
\label{fig:mask}
\end{figure}

\subsection{Ablation Studies}

We investigated which component had an impact on performance. We evaluated the classification performance by conducting ablation studies on model components and training steps. As shown in Table \ref{tab:result1}, our model without the prototype-based classifier (w/o prototype) and adaptive attention network (w/o $\bM$) exhibit degraded performance in all metrics by a large margin. Meanwhile, the performance of our model without intra-network relation encoding is less degraded. Inter-network relational learning appears to play a more critical role in encoding than intra-network learning. Step 2 is designed for counter-condition FC generation, therefore, it has less impact on classification.


As shown in Table \ref{tab:result2}, we compared the diagnostic performance depending on the factors that affect counter-condition classification. Our method demonstrated high classification performance in counter-condition classification. However, there was a decline in AUC and ACC when training without Step 2 and $\mathcal{L}_{\text{reg}}$ on both datasets. Specifically, there was a sharp decline in all metrics when training without Step 2. This indicates that the guidance provided by Step 2 is crucial for the successful generation of the simulated FC in the counter-condition. Without applying regularization, the SPC is sharply degraded on ABIDE dataset and there is an increase in the standard deviation of all metrics on the MDD datasets. This illustrates the influence of the residual class information in seed-based network features on the generation of counter-condition FC.

\begin{table}[]
\caption{Top five disease-related ROIs with large degree centrality (DC) from $\bM$. Note that these selected ROIs show significant group differences (*: $p$$<$0.05).}
\renewcommand{\arraystretch}{1.2}
\begin{tabular}{cccc}
\hline\hline
\multicolumn{4}{c}{\textbf{ABIDE}}                                                  \\ \hline
Craddock & Corresponding ROIs in AAL                           & ASD DC & TD DC \\ \hline
ROI 52*   & Cerebellum 9.L                                    & 0.356   & 0.488  \\
ROI 58*   & Precuneus.L/R, Vermis 4,5                        & 0.470   & 0.400  \\
ROI 78*   & Temporal Pole Mid/Sup.L                            & 0.310   & 0.510  \\
ROI 149*  & Frontal Sup Medial.L/R                             & 0.467   & 0.672  \\
ROI 185*  & Rolandic Oper.R, Temporal Sup.R                    & 0.382   & 0.555  \\ \hline\hline
\multicolumn{4}{c}{\textbf{REST-meta-MDD}}                                          \\ \hline
Craddock & Corresponding ROIs in AAL                           & MDD DC & HC DC \\ \hline
ROI 10*   & Cerebellum Crus1/6.R                   & 0.188   & 0.290  \\
ROI 41*   & Cerebellum 6.R, Cerebellum 4,5.R                      & 0.139   & 0.297  \\
ROI 72*   & Temporal Mid.L                                     & 0.158   & 0.262  \\
ROI 73*   & Precentral.L, Frontal Mid.L                        & 0.132   & 0.247  \\
ROI 140*  & Temporal Mid.R, Temporal Inf.R                     & 0.123   & 0.180  \\ \hline\hline
\end{tabular}
\label{tab:top5}
\end{table}

\section{Neuroscientific Analysis}
\label{sec:analysis}

\begin{figure*}[!ht]
\centerline{\includegraphics[width=0.9\textwidth]{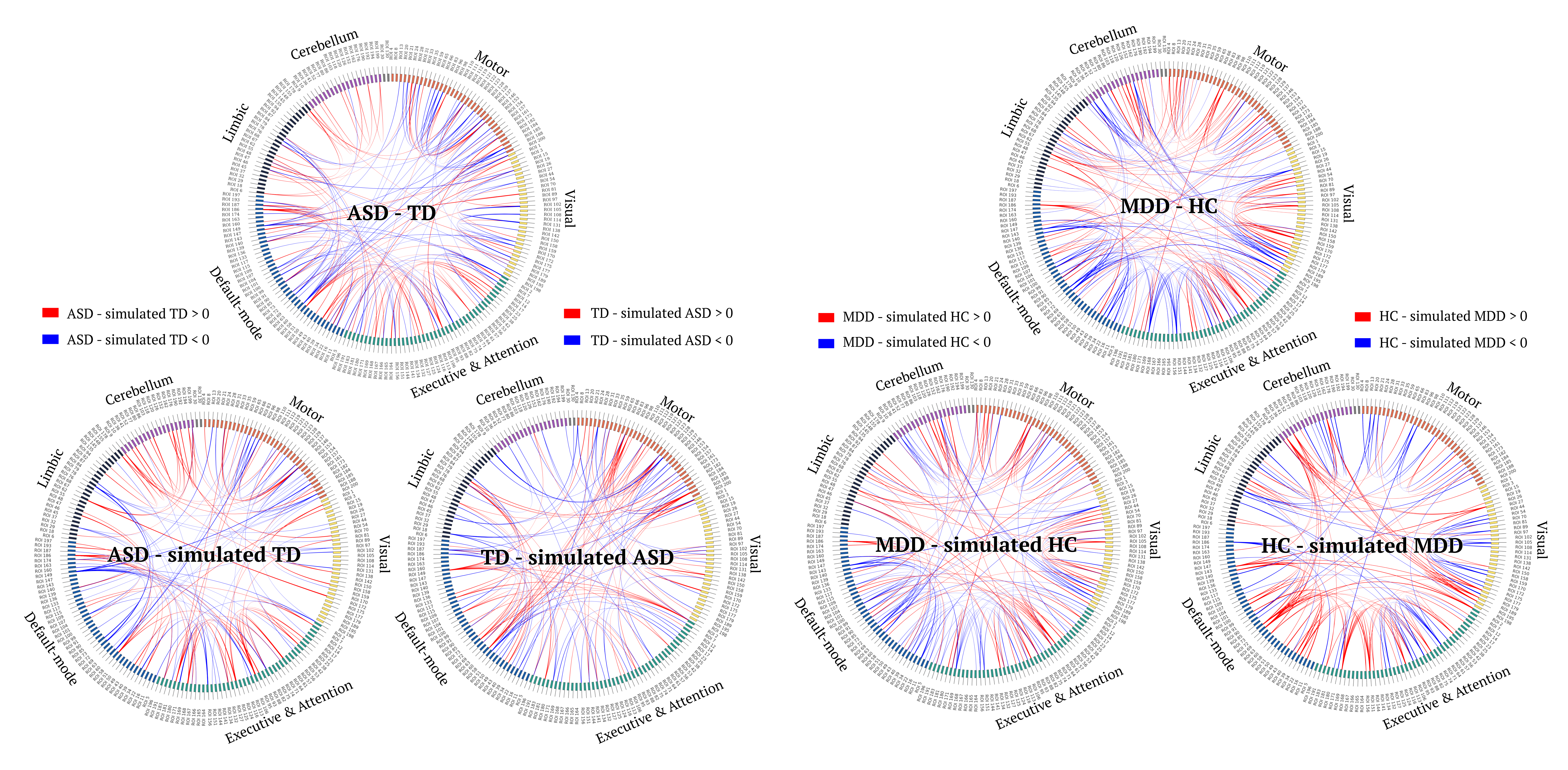}}
\caption{{\ES The average difference values between patients and healthy control for ABIDE and REST-meta-MDD datasets. The two upper circos plots illustrate the average changes between actual patients and actual normal controls. The four lower circos plots show the changes from patients to simulated healthy control (\ie, ASD-simulated TD and MDD-simulated HC), as well as changes from normal controls to simulated patients (\ie, TD-simulated ASD and HC-simulated MDD). The thickness of the lines describes the magnitude of the change. For clarity, please refer to the online version.}}
\label{fig:avgdiff}
\end{figure*}

\subsection{Feature Selection Analysis}
To demonstrate that our adaptive attention network generates an individual-specific mask $\bM$, we visualized the mean and standard deviation (STD) of the mask, as shown in Fig. \ref{fig:mask}. The average mask shows the connections that are attentive or irrelevant to the diagnosis. A higher average value indicates that the connection is more attentive. Meanwhile, the STD of the mask represents the connections that are consistently attentive or irrelevant across individuals. Connections with high average and low STD may be considered primary biomarkers associated with the disease. In contrast, connections with high average and high STD can be considered biomarkers associated with the individual characteristics of the disease. This selection of connections can provide valuable insights into the underlying mechanisms of the disease and facilitate the identification of potential biomarkers. {\ES It is worth mentioning that our method inherently provides explanations at the connection-level, elaborating on the disruptions to functional network properties in the analysis of brain disorders \cite{van2019cross}.} 

{\ES We further analyzed these changes of connections at the ROI-level as well. To figure out disease-related ROIs, we calculated the degree centrality of selected connections, which is one of the graph-theoretic concepts required for analyzing the brain connectivity architecture in fMRI studies \cite{wang2010graph}. Degree centrality in the context of FC provides a way to understand the importance of brain regions as \textit{hubs} that are highly connected to other regions within the overall network. Subsequently, we performed a group t-test with a significance level of $p$$<$0.05. We then identified the top five disease-related ROIs that exhibited the highest degree centrality among those that showed statistically significant differences in Table \ref{tab:top5}.}

\subsection{Counter-condition Analysis}
For quantitative analysis, we considered all individuals in the training, validation, and test datasets and filtered the correctly predicted subjects in both conventional disease and counter-condition classification. For the counter-condition explanations, we assumed two scenarios: (1) What if the patient becomes normal ($\hat{\bX}^{\text{control}}$)? (2) What if a healthy is diagnosed with disease ($\hat{\bX}^{\text{patient}}$)? As explained in Sec. \ref{sec:decforcountercondition}, our decoder generates FC from seed-based network features and prototypes to simulate the mechanisms by which functional connections are changed depending on the prototypes of the counter-condition. To calculate the amount of change, we subtracted the counter-condition FC from the reconstructed FC with its own class information. Then, we excluded the extreme 1\% connections (\ie, 199 connections for Craddock 200 atlas based on the absolute values for each individual. 
As a general understanding of the disease, the average changes in functional connectivity are displayed in Fig. \ref{fig:avgdiff}. {\ES The connection patterns of ASD-simulated TD and TD-simulated ASD show opposite tendencies to each other. This suggests that to change from a patient to a healthy state, or vice versa, the disease-related connections should be modified in reverse. Additionally, we calculated and added the actual average differences between patients and healthy controls. The patterns between ASD-TD and ASD-simulated TD are alike, indicating that the simulated FC properly represents the functional connectivity of the opposite condition.} {\ES In line with previous research \cite{nomi2015developmental, liang2020biotypes}, we also identified patterns of both functional hypo- and hyper-connectivity in individuals with ASD.} To deal with differences in neuropathological abnormalities, we focused on the patients and their counter-condition FC $\hat{\bX}^{\text{control}}$ for afterwards analyses. 



\subsection{Subtype Analysis}

\begin{figure}[t!]
\centering
\includegraphics[width=0.45\textwidth]{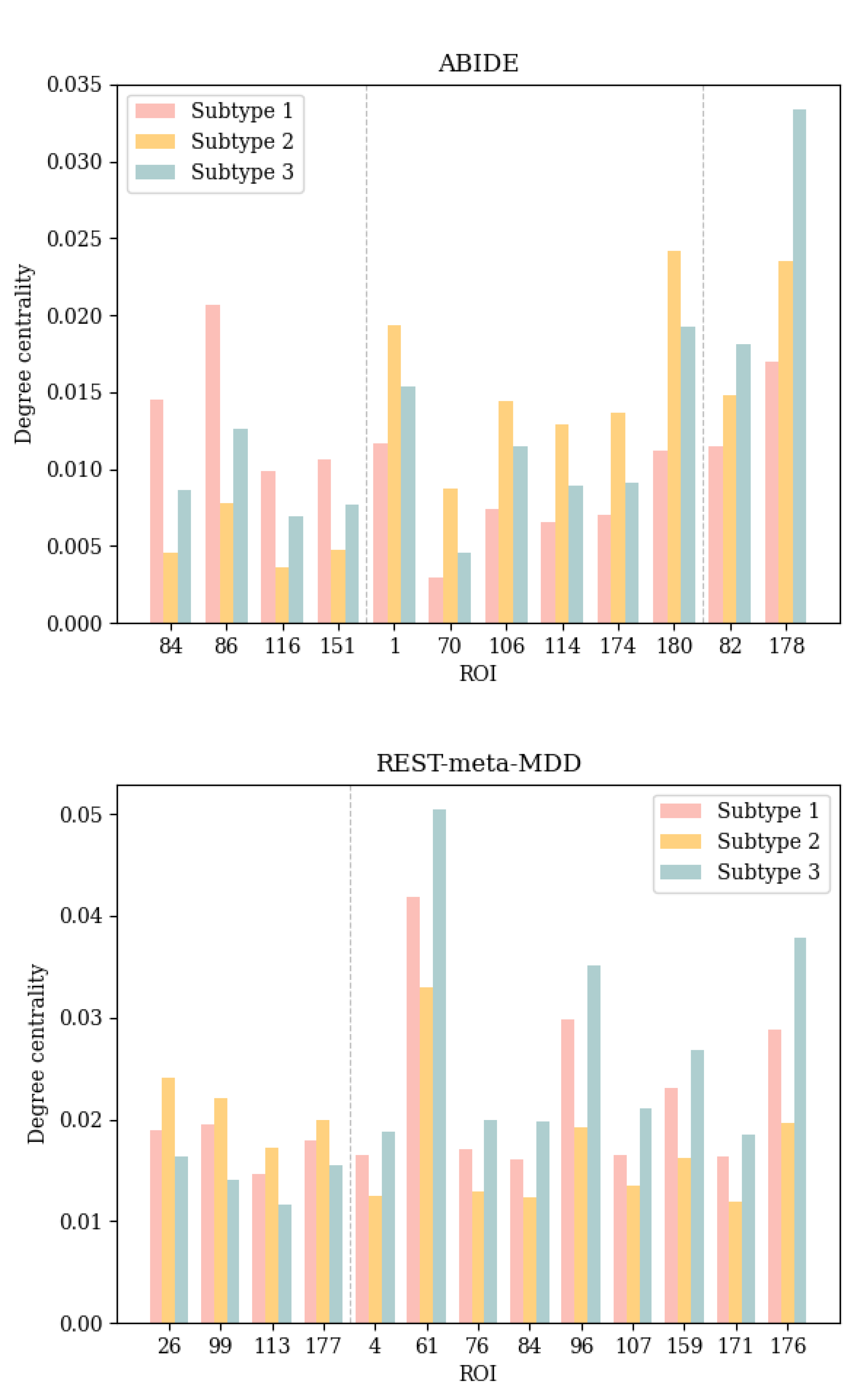}
\caption{The average degree centrality of significant ROIs for three subtypes of MDD and ASD ($p<$0.05).}
\label{fig:subtype_roi}
\end{figure}

\begin{figure}[t!]
\centering
\includegraphics[width=0.45\textwidth]{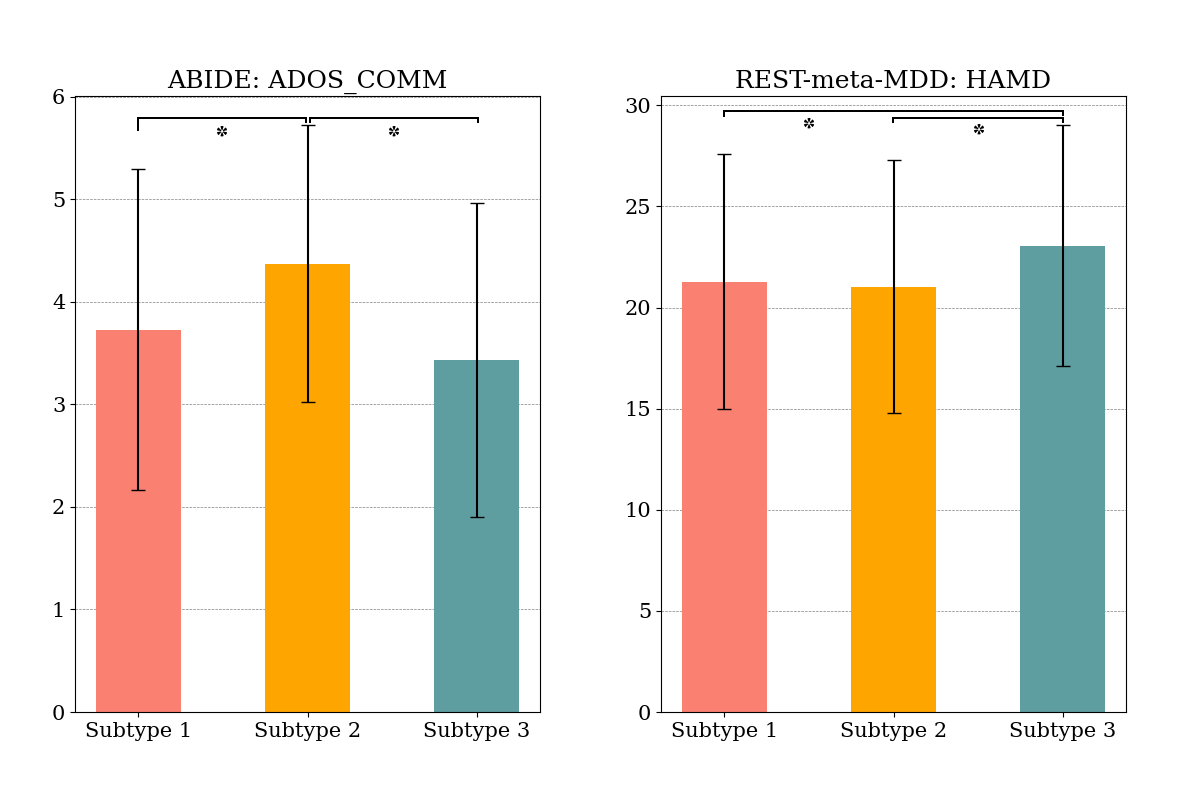}
\caption{The average ADOS-COMM scores of ASD patients and HAMD scores of MDD patients for three subtypes. *: statistical significance}
\label{fig:score}
\end{figure}

To deal with the diversity of malfunctioning connections, we conducted a clustering analysis of the changes in functional activities in ASD and MDD patients, \ie, differences between patients' FC and their counter-condition FC. We categorized the patients into subtypes using a Ward linkage-based hierarchical clustering method, widely used in neuroimaging studies \cite{lee2021unified}. {\ES According to previous research that suggested 2$\sim$5 subtypes of disease \cite{van2012data, spencer2022using}, we empirically set the number of clusters as three. 

To analyze neuroscientific differences among subtypes, we found subtype-related ROIs with these functional changes. First, we calculated the degree centrality of selected connections, which is one of the graph-theoretic concepts required for analyzing the brain connectivity architecture in fMRI studies \cite{wang2010graph, lin2022abnormal}. Then, we conducted ANOVA ($p$$<$0.05) with Bonferroni correction. We reported subtype-related ROIs in Fig. \ref{fig:subtype_roi}.

As depicted in Fig. \ref{fig:score}, we also observed that the ADOS communication (COMM) score that measures patients' verbal and non-verbal communication skills was significantly related ($p<$0.05) with subtypes. Specifically, patients belonging to Subtype 2 exhibited higher communication scores compared to those in the other subtypes. For MDD patients, we utilized the Hamilton Depression Rating Scale (HAMD) and found that the three identified subgroups exhibited different average HAMD scores. Subtype 3 shows the highest-averaged
score.}

\section{Conclusion}
\label{sec:conclution}
In this study, we proposed a unified framework that integrates diagnosis and neuroscientific explanation. Our framework has three main advantages: no pre-selected features by the use of adaptive attention networks, no pre-defined ROI relations through the utilization of functional network relation encoder, and no additional training for the explanation thanks to counter-condition learning. In our experiments on the ABIDE and REST-meta-MDD datasets, the proposed method outperformed competing methods. We believe that our counter-condition analysis could facilitate precise diagnoses and effective treatments for the patients.

\appendices

\bibliographystyle{IEEEtran} 
\bibliography{tmi.bib}

\end{document}